\def\year{2020}\relax
\documentclass[letterpaper]{article} % DO NOT CHANGE THIS
\usepackage{aaai20}  % DO NOT CHANGE THIS
\usepackage{times}  % DO NOT CHANGE THIS
\usepackage{helvet} % DO NOT CHANGE THIS
\usepackage{courier}  % DO NOT CHANGE THIS
\usepackage[hyphens]{url}  % DO NOT CHANGE THIS
\usepackage{graphicx} % DO NOT CHANGE THIS
\usepackage{graphicx}  % DO NOT CHANGE THIS
\usepackage{epsfig}
\usepackage{tabularx}
\usepackage{amsmath,mathtools}
\usepackage{amssymb}
\usepackage[inline]{enumitem}
\usepackage[dvipsnames]{xcolor}

\usepackage{booktabs}
\usepackage{tabularx}
\usepackage{multirow}
\usepackage{colortbl}
\usepackage{subcaption}
\usepackage{float}

\usepackage{algpseudocode,algorithm,algorithmicx}
\usepackage{xspace}
\usepackage{interval}
\usepackage{bm,upgreek}

\usepackage{xcolor}

\def\HiLi{\leavevmode\rlap{\hbox to 0.85\linewidth{\color{gray!20}\leaders\hrule height .8\baselineskip depth .5ex\hfill}}}

\makeatletter
\newcommand{\algcolor}[2]{%
  \hskip-\ALG@thistlm\colorbox{#1}{\parbox{\dimexpr\linewidth-2\fboxsep}{\hskip\ALG@thistlm\relax #2}}%
}

\makeatother

\makeatletter

\newcommand*\wthelper[2]{%
        \hbox{\dimen@\accentfontxheight#1%
                \accentfontxheight#11.3\dimen@
                $\m@th#1\widetilde{#2}$%
                \accentfontxheight#1\dimen@
        }%
}

\newcommand*\accentfontxheight[1]{%
        \fontdimen5\ifx#1\displaystyle
                \textfont
        \else\ifx#1\textstyle
                \textfont
        \else\ifx#1\scriptstyle
                \scriptfont
        \else
                \scriptscriptfont
        \fi\fi\fi3
}
\makeatother

\newcommand{\instagram}{{\scshape ImageNet-Instagram}\xspace}
\newcommand{\imagenet}{{\scshape ImageNet}\xspace}

\newcommand{\system}{DS\xspace}
\newcommand{\gsystem}{gDS\xspace}

\definecolor{Gray}{gray}{0.85}
\newcolumntype{g}{>{\columncolor{Gray}} c}

\newcommand{\xx}{{\bm x}}
\newcommand{\yy}{{\bm y}}

\newcommand{\vv}{{\bm v}}
\newcommand{\zz}{{\bm z}}

\title{Recognizing Instagram Filtered Images with Feature De-stylization}
\author{Zhe Wu\textsuperscript{\rm 1}\thanks{This work was done when the author was at Univ. of Maryland.},
Zuxuan Wu\textsuperscript{\rm 2}\thanks{Corresponding author.},
Bharat Singh\textsuperscript{\rm 2},
Larry S. Davis\textsuperscript{\rm 2}\\ 
\textsuperscript{\rm 1}Comcast Applied AI Research, \textsuperscript{\rm 2}University of Maryland,College Park\\ %If you have multiple authors and multiple affiliations
zhe\_wu@comcast.com, \{zxwu, bharat, lsd\}@cs.umd.edu% email address must be in roman text type, not monospace or sans serif
}
\begin{document}

\maketitle

\begin{abstract}
Deep neural networks have been shown to suffer from poor generalization when small perturbations are added (like Gaussian noise), yet little work has been done to evaluate their robustness to more natural image transformations like photo filters. This paper presents a study on how popular pretrained models are affected by commonly used Instagram filters. To this end, we introduce ImageNet-Instagram, a filtered version of ImageNet, where 20 popular Instagram filters are applied to each image in ImageNet. Our analysis suggests that simple structure preserving filters which only alter the global appearance of an image can lead to large differences in the convolutional feature space. To improve generalization, we introduce a lightweight de-stylization module that predicts parameters used for scaling and shifting feature maps to ``undo'' the changes incurred by filters, inverting the process of style transfer tasks. We further demonstrate the module can be readily plugged into modern CNN architectures together with skip connections. We conduct extensive studies on ImageNet-Instagram, and show quantitatively and qualitatively, that the proposed module, among other things, can effectively improve generalization by simply learning normalization parameters without retraining the entire network, thus recovering the alterations in the feature space caused by the filters. 
\end{abstract}

%%%%%%%%% BODY TEXT
% !TEX ROOT=./main.tex
\section{Introduction}
Convolutional neural networks (CNNs) demonstrate impressive recognition accuracies on standard benchmarks like \imagenet~\cite{deng2009imagenet}, even surpassing human-level performance~\cite{he2015delving} especially when recognizing fine-grained objects; and as a result, these trained models are widely applied to a variety of applications in real-world. However, recent studies have shown CNNs are vulnerable to even small perturbations~\cite{hendrycks2018benchmarking} like a Gaussian noise or blur, resulting in significant performance degradation, let alone maliciously injected adversarial noises~\cite{goodfellow2014generative,kurakin2016adversarial}. This raises doubts about whether these models can offer reliable recognition performance, particularly when evaluated on images from social media, where images are generally modified (filtered) extensively by adjusting parameters like brightness, color, and contrast or their combinations. For example, Instagram provides 40 pre-defined filters and users can apply these filters to make photos look appealing with only a few clicks. These filtered images contain various visual effects, incurring far more complicated perturbations compared to Gaussian noises~\cite{hendrycks2018benchmarking,geirhos2018generalisation}. Thus, models pretrained on natural images may fail when tested on these filtered images.

\begin{figure}[t!]
\includegraphics[width=1.0\linewidth]{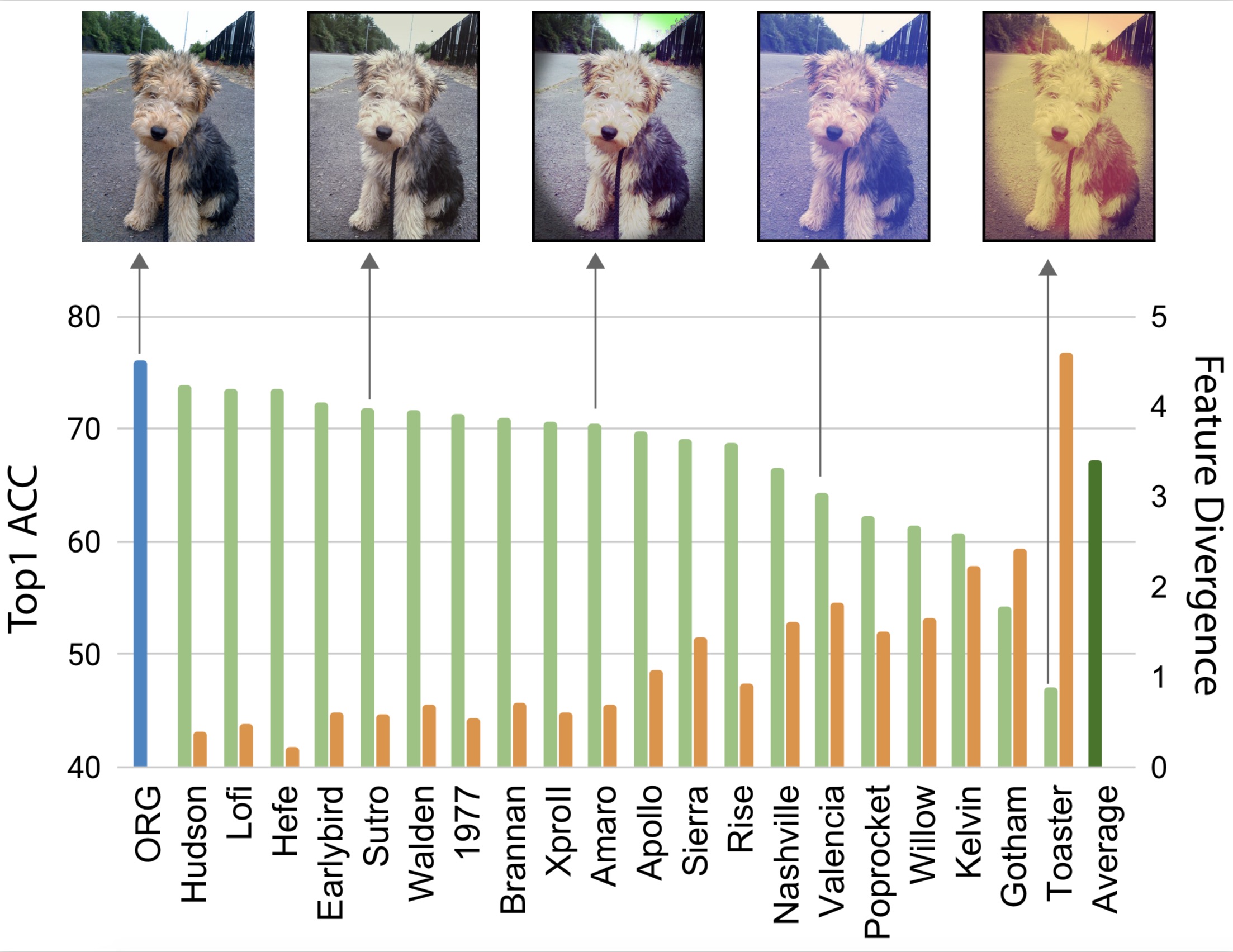}
\label{fig:raise}
\vspace{-10pt}
\caption{Performance of ResNet50 pretrained on \imagenet (Green) and \texttt{Conv5} feature divergence (Orange).  X-axis: Different types of Instagram filters. Left-Y-axis: Top1 accuracy; right-Y-axis: Feature Divergence. }
\label{fig:teaser}
\end{figure}

Extensive studies have been conducted to improve the generalization~\cite{zhang2017understanding} of deep neural networks and techniques like Dropout~\cite{srivastava2014dropout} and BatchNorm~\cite{ioffe2015batch} can effectively reduce overfitting.  The generalization ability of deep networks is usually measured on a held-out testing set; or related tasks using features that are finetuned to be task-specific~\cite{kornblith2018better} or explicitly adapted for distribution alignment~\cite{DBLP:conf/cvpr/TzengHSD17}. In this paper, we estimate generalization from a different perspective---by testing on filtered images whose appearance is significantly modified but the structure and semantics are preserved. Such filtered images are prevalent on social media and it is critical to develop networks that generalize well on these images.

In light of this, we systematically study the robustness of modern CNN architectures to widely used Instagram filters for image classification, and introduce a simple yet effective approach that helps the generalization of these architectures on filtered images .
%by learning a network with $\sim95\%$ less parameters
To begin with, we create a new dataset, referred to as \instagram, which contains images transformed from \imagenet using 20 common Instagram filters---each original image in \imagenet is applied with these 20 filters, generating 20 copies that share the same semantic content but differ in appearance (See Fig.~\ref{fig:dataset} for an example). We then analyze the performance of several top-performing CNNs on the newly constructed dataset.  The results suggest that
dramatic changes in appearance lead to huge feature differences compared to those original images, which further result in significant performance drops (cf. ``Toaster'' and ``Sutro''  in Fig.~\ref{fig:teaser}).

Therefore, we posit that the visual effect brought by filters not only changes the style of original images but also injects style information into feature maps, resulting in shifts from original feature representations. If we can find a way to remove style information in these feature maps, they will be closer to those of the original samples. This is essentially the inverse process of style transfer tasks that aim to add style information into features, typically done with instance normalization (IN)~\cite{ulyanov2016instance} to scale and shift feature maps at each channel. Then the question becomes, can we learn a set of parameters that re-normalize feature maps with IN by scaling and shifting to ``undo'' the changes caused by filters. If so, the performance on filtered images can be improved by simply tuning these parameters to produce re-normalized feature maps. 
% This is appealing as the number of floating point operations (FLOPs) needed to learn the parameters for normalization are far more less than finetuning the weights in the entire network. 

% \todo{How? this still requires doing back-prop for the entire network if I am not wrong...I would try to sell this as an interesting observation, yeah makes sense! }

To this end, we propose a lightweight de-stylization module (DS), which contains a five-layer fully-connected network operating on feature maps encoded by a pretrained VGG encoder. The DS module outputs multiple sets of parameters, each of which is used to scale and shift feature maps in a corresponding IN layer of a base network. The DS module can be readily used in networks like IBN~\cite{pan2018two} where IN layers are used for recognition tasks. To further extend the DS module to modern architectures without IN layers, we introduce a generalized DS (\gsystem), which performs IN with the de-stylization module on feature maps in modern networks but the normalized feature maps are further shortcut by skip connections. Such a design ensures style information in feature maps caused by filters can be removed with learned normalization parameters without destroying the optimized feature maps in the base network. We conduct extensive results on the newly proposed \instagram dataset, and we demonstrate that both \system and \gsystem can effectively improve generalization when applying pretrained models on filtered images by simply learning normalization parameters without retraining the whole network and \gsystem is compatible with modern CNN architectures even without IN layers. Our qualitative results also suggest that \gsystem can indeed transform features of filtered image to be similar to those before filtering.

%!TEX ROOT=main.tex
\section{Related Work}
\textbf{Corruption Robustness}. There are a few recent studies investigating the robustness of deep neural networks to corrupted or noisy inputs~\cite{hendrycks2018benchmarking,geirhos2018generalisation}. 
Heydrycks \textit{et al.} \cite{hendrycks2018benchmarking} introduce two variants of \imagenet to benchmark the robustness of deep models to common corruptions and perturbations. 
The results suggest that deep models demonstrate instabilities to even small changes in input distributions. 
Geirhos \textit{et al.} \cite{geirhos2018generalisation} study the robustness of humans and several CNNs on different types of degraded images. Our work differs from these methods as we focus on filtered images that contain a series of sophisticated and carefully designed transformations as opposed to basic transformations like Gaussian noise and rotations in~\cite{hendrycks2018benchmarking,geirhos2018generalisation}. 
In addition, filtered images are created intentionally to make images aesthetically pleasing and thus improving generalization on these filtered images enjoys wider applications. 
% Chen \textit{et al.} explore the performance of classification CNNS on Instagram filters on a small subset of images in \imagenet while we focus on the entire \imagenet and we also delve deeper into feature space to analyze changes caused by filters~\cite{chen2015filter}.

\textbf{Domain Adaptation}. Our work is also related to domain adaptation, or transfer learning, which aims to adapt a learned model to new tasks by aligning the feature distributions of the source task with that of the target task. Existing approaches usually minimize distances such as Maximum Mean Discrepancy (MMD)~\cite{saito2018maximum}, first-order and second order statistics~\cite{DBLP:conf/eccv/SunS16} or make features indistinguishable with adversarial loss functions~\cite{ganin2016domain,DBLP:conf/cvpr/TzengHSD17}. However, these methods require training or finetuning the majority of weights in networks, which is computationally expensive particularly when using adversarial loss functions~\cite{arjovsky2017towards}. In contrast, we focus on how to improve generalization with a lightweight module. Recently, Li \textit{et al.}~\cite{li2016revisiting} introduce Adaptive Batch Normalization (AdaBN) that applies a trained network to a target domain without changing the model weights. AdaBN collects the statistics of Batch Normalization layer on target domains before final testing, and use the target domain statistics for normalization. However, it requires the distribution of test samples before testing, and is designed for adaptation to a unique domain. Our proposed approach, on the other hand, does not assume access to testing data beforehand (only access to the filtering function is needed), and performs normalization based on the appearance of the testing sample and thus can be applied to a mixture of different domains at the same time. 

\textbf{Image Synthesis for De-stylization}. Recent advances in image-to-image translation provide a way to remove filtering effect such that de-stylized images can be directly input into the original model learned on natural images. In particular, image-to-image translation aims to generate images from a source domain in the style of a target domain. This can be achieved using generative models that enforce cycle-consistency~\cite{DBLP:conf/iccv/ZhuPIE17,huang2018munit} or neural style transfer algorithms~\cite{gatys2016image,huang2017adain}. To remove the filtering effect with generative models is hard as adversarial loss functions are difficult to optimize and it requires modeling a many-to-one mapping if there are multiple source domains. De-stylization with style transfer is also challenging since it is difficult to select  images from the source as references. In addition, both methods require training an additional large model to generate images while our approach aligns features maps with a lightweight network.

\textbf{Feature Normalization.} 
Feature normalization is an essential component in modern deep CNNs. Batch Normalization \cite{ioffe2015batch} is widely used for faster convergence and better performance. Instance Normalization \cite{ulyanov2016instance} helps achieve good image stylization performance, as the channel-wise feature statistics are shown to contain sufficient style information~\cite{ulyanov2016texture}. In contrast to these methods operating on feature maps, there are some studies on conditional normalization which modulates feature maps with additional information. Conditional Batch Normalization \cite{dumoulin2016learned} and Adaptive Instance Normalization (AdaIN) \cite{huang2017adain} adjust the normalization layer parameters based on external data inputs, and thus are able to achieve image stylization on a diverse set of styles. However, these normalization methods are mainly designed for generative tasks, and have not been used in discriminative models for recognition. In our work, we demonstrate the proposed de-stylization module is applicable to several modern deep networks to improve their generalization on filtered images for recognition.

\section{Imagenet-Instagram}
\label{sec: dataset}

\begin{figure*}[ht!]
\includegraphics[width=\linewidth]{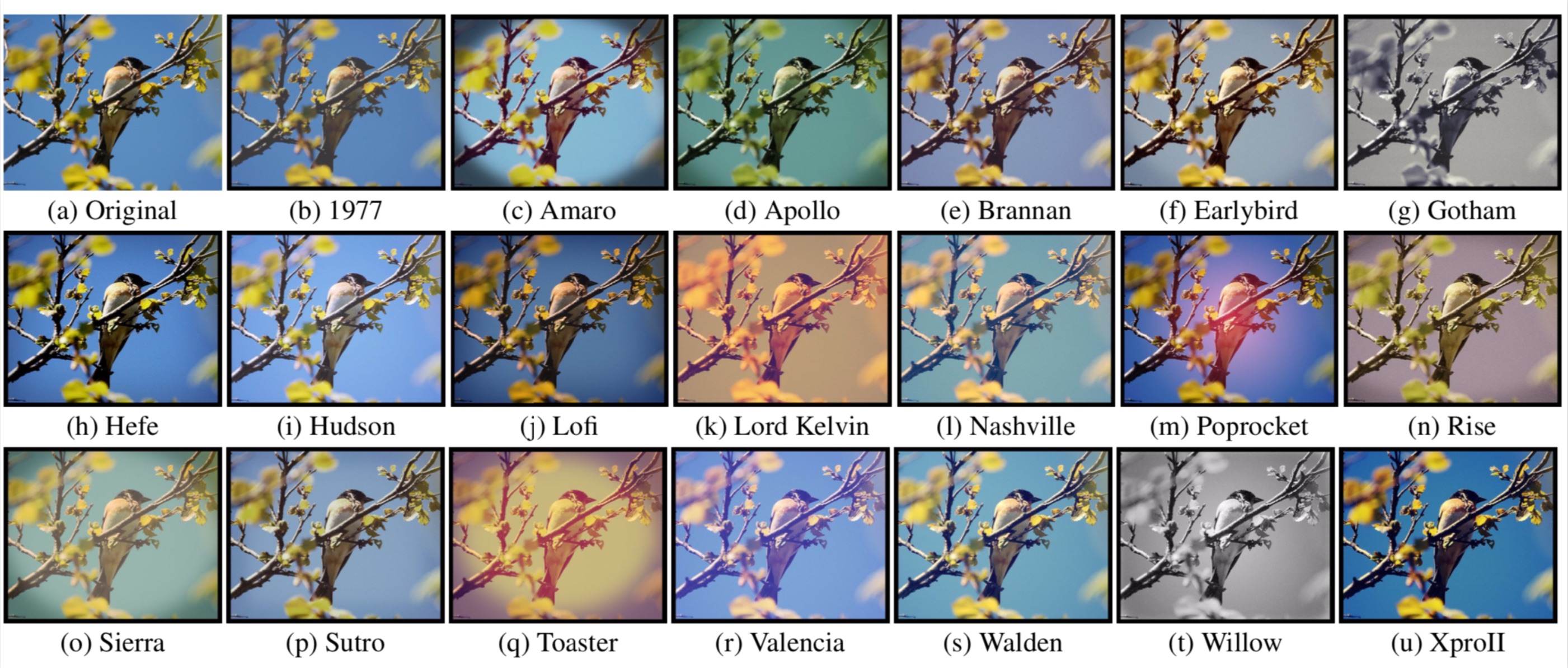}
\vspace{-15pt}
\caption{A sampled image from \imagenet and its 20 Instagram filtered versions.}
\label{fig:dataset}
\end{figure*}

Many social media apps provide a variety of artistic image filters to help users editing the photos. For example, Instagram has 40 pre-defined photo filters. These filters are combinations of different effects such as curve profiles, blending modes, color hues, \textit{etc.}, which make the filtered photos aesthetically appealing. We select 20 commonly used Instagram filters and apply them to each image in \imagenet, the resulting new dataset is named as \instagram.
Figure.~\ref{fig:dataset} illustrates one sampled image from \imagenet and its 20 filtered versions with Instagram filters. As we can see, different Instagram filters generate different photo styles. For example, ``Amaro'' filter brightens the center of the image and adds vignetting to the border of the image. Some filters like ``1977'' and ``Hefe'' adjust the contrast of the image slightly without creating dramatic effects, while other filters like ``Gotham'' and ``Willow'', discard some important information like color.

To evaluate the performance of modern CNN architectures on these filtered images, we run a ResNet50~\cite{he2016deep} pretrained from \imagenet on the validation set of \instagram directly. The results are shown in Figure~\ref{fig:teaser}.  We observe that the average Top-1 accuracy drops from 76.13\% on  \imagenet validation set to 67.22\% on \instagram. Filters like ``Gotham'' and ``Toaster''  create significantly different appearances at the same time suffer from 
drastic performance drop (\textit{eg.}, $21.13$ and $29.03$ absolute percentage points for ``Gotham'' and ``Toaster'' respectively). It is also surprising to see the performance of filters like ``1977'' which bring slight differences in color also drops by $5\%$.

To better understand why pre-trained ResNet50 suffers from poor performance on \instagram, we analyze the feature divergence~\cite{pan2018two} (see supplementary material for definition) of \imagenet and \instagram samples. 
Specifically, we compute features from the \texttt{Conv5} layer for images in both \imagenet and \instagram. For each filter type, we compute the feature divergence of \texttt{Conv5} between \imagenet and \instagram on validation set. Figure.~\ref{fig:teaser} presents the results. 
We can clearly see the correlations between feature divergence and the performance on the validation set of \instagram---large feature divergence translate to lower accuracies (see ``Toaster'', ``Gotham'' and ``Lord Kelvin''). 

% We follow the definition of feature divergence metric in~\cite{li2016revisiting}. Denote the feature from \imagenet as $F_a$ and feature from \instagram as $F_b$, the feature divergence is defined using symmetric KL divergence:
% \begin{equation}
% \label{eqn:div}
% D(F_a||F_b)= \textrm{KL}(F_a||F_b) + \textrm{KL}(F_b||F_a)
% \end{equation}
% If we assume $F_a$ and $F_b$ follow Gaussian distribution with $(\mu_a, \sigma_a^2)$ and $(\mu_b,\sigma_b^2)$, then KL divergence can be expressed as:
% \begin{equation}
% \label{eqn:div}
% \textrm{KL}(F_a||F_b) = \log\frac{\sigma_b}{\sigma_a}+\frac{\sigma_a^2+(\mu_a-\mu_b)^2}{2\sigma_b^2}-\frac{1}{2}.
% \end{equation}

%!TEX root=main.tex
\section{Method}
Since feature divergence is positively correlated with performance drop, it would be ideal to reduce such mismatch induced by applying filters---the removal of style information encoded in feature maps. This is similar in spirit to style transfer tasks but in the reversed direction; style transfer approaches incorporate style information with instance transformation using a set of affine parameters that are either learned~\cite{ulyanov2016instance} or computed from another image~\cite{huang2017adain}. Such an operation simply scales and shifts feature maps to add style information, which thus motivates us to invert this process in order to remove the style information. To this end, we introduce a lightweight de-stylization module (DS), which predicts a set of parameters used to normalize feature maps and hence further remove encoded style information therein. Then we discuss how to easily plug the module into modern architectures such that generalization of these networks can be improved on filtered images.

\textbf{A lightweight de-stylization module (DS)}. As mentioned earlier, style transfer pipelines usually rely on instance normalization (IN) to normalize features ~\cite{ulyanov2016instance,huang2017adain}. In particular, IN normalizes features per channel for each sample separately using mean and variance computed in each channel. Denote the $i$th channel of feature map $\xx$ as $\xx_i$, the mean and variance of $\xx_i$ as $\mu(\xx_i)$ and $\sigma^2(\xx_i)$, then the {IN} operation is defined as: 
\begin{equation}
\yy_i = \gamma_i\cdot\frac{\xx_i-\mu(\xx_i)}{\sigma(\xx_i)}+\beta_i
\end{equation}
%% Batch Normalization
\begin{equation}
\yy_i = \gamma_i\cdot\frac{\xx_i-\mu_B}{\sigma_B}+\beta_i
\end{equation}
where $\gamma_i$ and $\beta_i$ are affine parameters for channel $i$. By learning a set of affine paramters $\gamma$ and $\beta$, IN facilitates the transfer of the original image to a different style for image synthesis. A recently proposed IBN~\cite{pan2018two} further demonstrates IN layers can be used in discriminative tasks like classification, by simply replacing half of BN layers in a ResNet BottleNeck with IN layers, and demonstrates good generalization across different domains.

Filtering a realistic image with an Instagram filter changes its intermediate feature maps when passing it through the CNN, and such changes will further lead to significant feature divergence as shown in Fig.~\ref{fig:teaser}. Given that IN layers can encode style information into feature maps with affine transformations, a natural question to ask is: Can we simply finetune the IN layers to obtain a different set of affine parameters that are able to remove style information in feature maps caused by applied filters? This allows a network to quickly generalize to filtered images without re-training the entire network. However, finetuning IN parameters indicates the same set of affine parameters of each channel are shared by all images, which might be viable if we are targeting at a single type of filter rather than 20 different filters.

Recall our goal is to tune IN parameters that are tailored for each filter, such that for each type of filtered image, the IN operation can successfully undo the changes in feature maps caused by input perturbations. This is the inverse process of \emph{arbitrary} style transfer, where the feature maps from a source image is normalized using different affine parameters based on the styles to be transferred. Thus, we build upon adaptive instance normalization (AdaIn), which enables the transfer of the style of an arbitrary image to a source image~\cite{huang2017adain}. Formally, \textrm{AdaIN} is defined as follows:

\begin{equation}
\label{eqn:adain}
	\textrm{AdaIN}(\xx_i,\zz_{m}, \zz_{v}) = \zz_{m,i}\frac{\xx_i-\mu(\xx_i)}{\sigma(\xx_i)} + \zz_{v,i}	\textrm{,}	%
\end{equation}
where each feature map $\xx_i$ is normalized separately, and then scaled and shifted using the corresponding mean $\zz_{m,i}$ and variance $\zz_{v,i}$ of feature maps $\zz$ of a target image.  In style transfer tasks~\cite{huang2017adain}, feature maps $\zz$ are computed with a fixed VGG encoder and used for normalization for only once before generating an image. In contrast, we wish to adaptively normalize all Instance Normalization layers in a network to fully recover changes caused by filters. 

\begin{figure}[t!]
 \centering
 \includegraphics[width=0.7\linewidth]{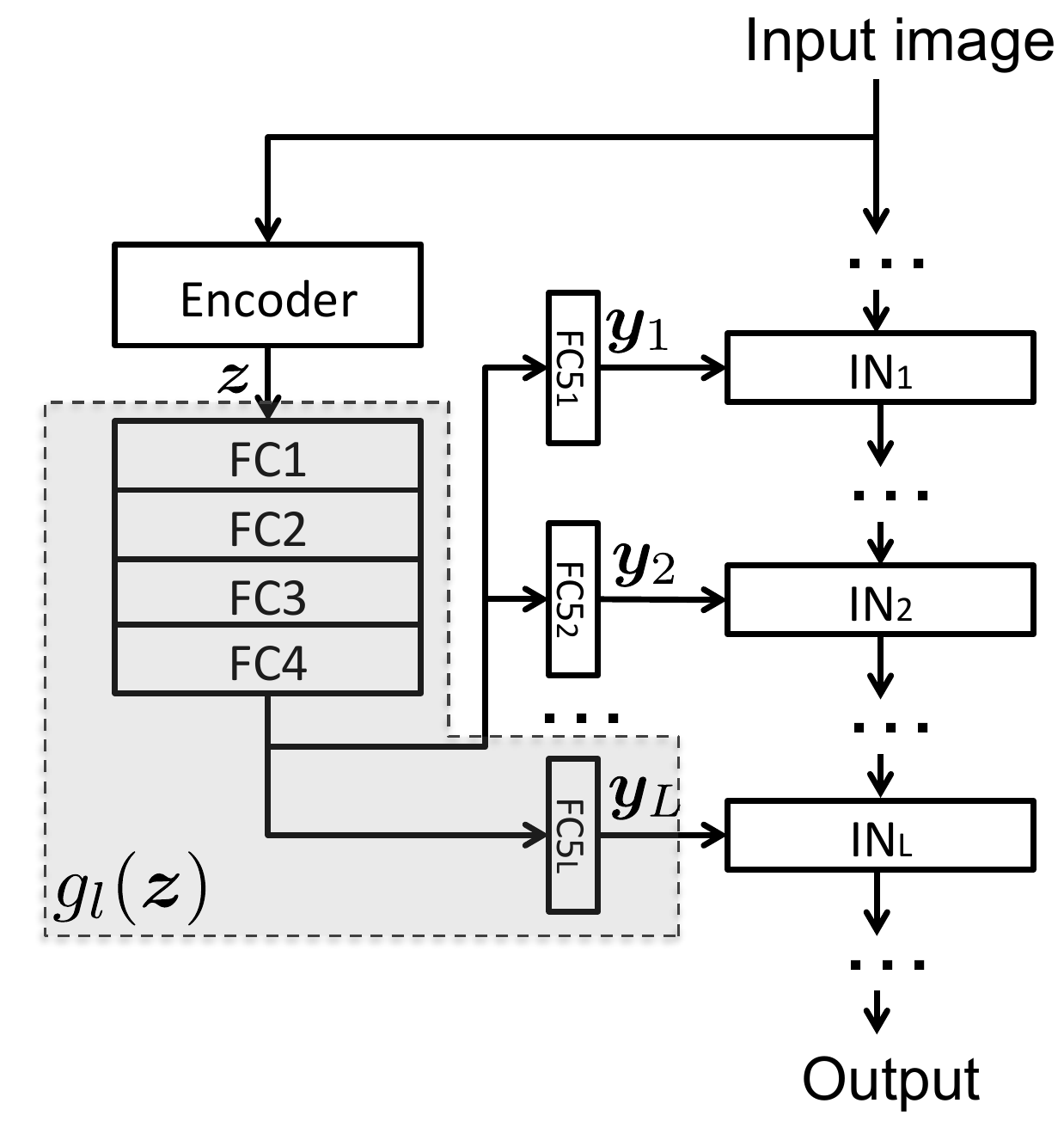}
\vspace{-5pt}
\caption{Our proposed framework, with DS module highlighted.}
 \label{fig:framework}
 \vspace{-10pt}
\end{figure}

To this end, we introduce a lightweight de-stylization module, which generates the affine parameters used for instance normalization in all IN layers. In particular, suppose there are $L$ IN layers in a network, we use a five-layer fully connected network to map $\zz$, the feature maps encoded by an VGG encoder, denoted as $H(\cdot)$, to $L$ vectors $\{\yy_l\}_{l=1}^{L}$, where each vector $\yy_l\in \mathbb{R}^{2C_l}$ is used to normalize the corresponding IN layer of $C_l$ channels with Eqn.~\ref{eqn:adain}. This is achieved by splitting $\yy_l$ into two parts $\yy_l = [\yy_{lm}, \yy_{lv}]$, with $\yy_{lm}$ and $\yy_{lv}$ denoting the predicted mean and variance for normalization. 
The first four fully-connected layers are used for transformation, denoted as $\mathcal{F(\cdot)}$. The final layer contains $L$ heads, denoted as  $\{f_l(\cdot)\}_{l=1}^L$, with each head corresponding to one of the $L$ IN layers in the network. 
Formally, we define de-stylization module for layer $l$ as $g_l(\cdot)$, then:
\begin{equation}
\label{eqn:head}
\yy_l = g_l(\zz) = f_l(\mathcal{F}(\zz))
\end{equation}
where $\zz=H(I)$, $I$ is the input Image. The DS module is illustrated in Fig.~\ref{fig:framework}.

% denoted as $f$, in which the first four fully-connected layers are used for transformation. And the final layer contains $L$ heads, with each head corresponding to one of the $L$ IN layers in the network. Each head computes two vectors, representing the parameters used for normalization of the corresponding IN layer. More formally, the outputs of the network is defined as:
% \begin{equation} 
% \label{eqn:gen}
% \hat{\zz} = f(\zz) \in \mathcal{R}^{L\times 2C_L},
% \end{equation}
% where $C_L$ denotes the number of channels in layer $L$. \todo{describe this part}

\begin{figure}[t!]
 \centering
 \includegraphics[width=0.8\linewidth]{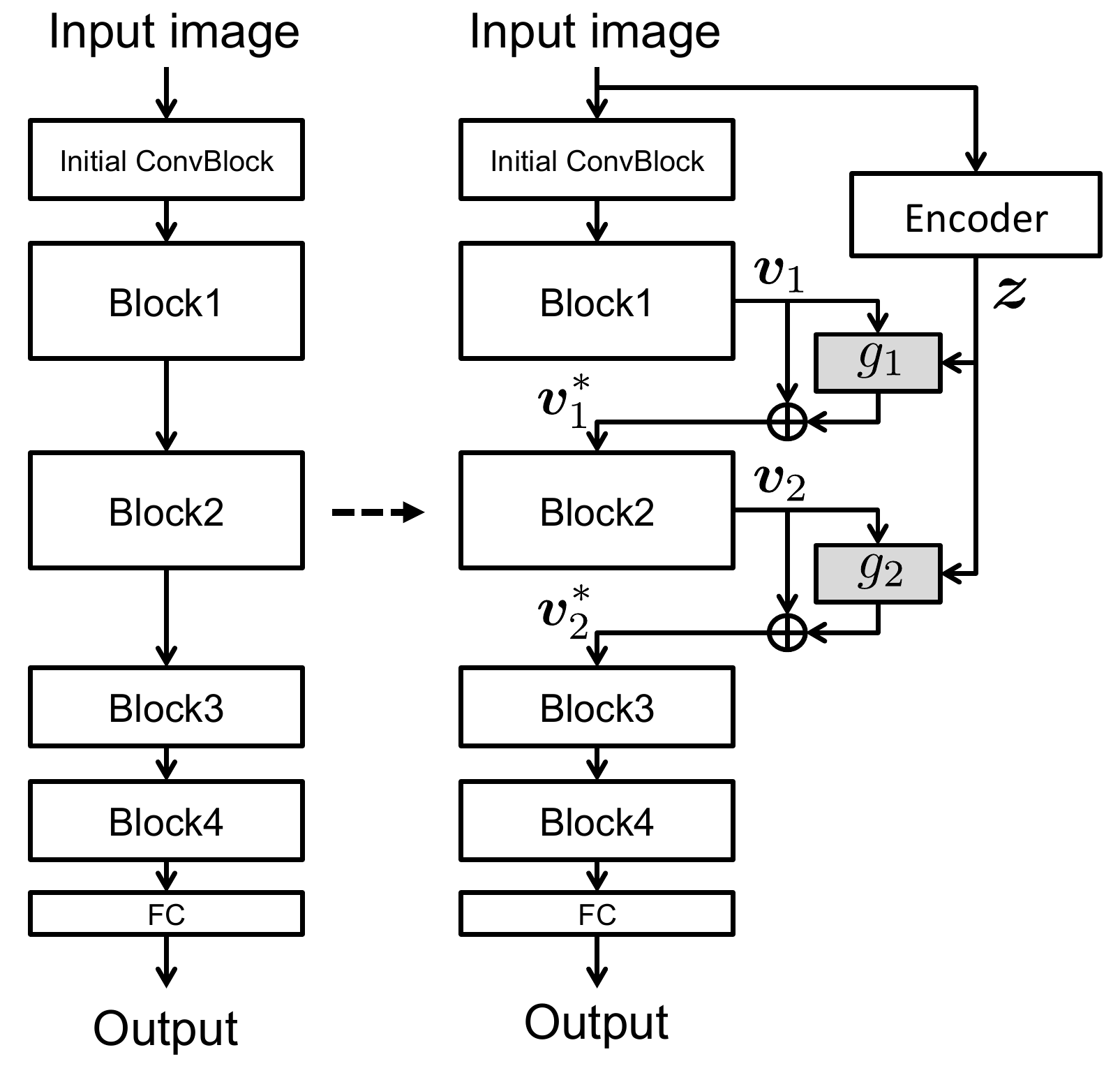}
 \vspace{-5pt}
\caption{Generalized de-stylization. Left: Orginal 5-block CNN prototype. Right: \gsystem applied on Block1 and Block2 of the CNN.}
 \label{fig:genIN}
\end{figure}

\textbf{Extension to modern architectures}.
We are now able to generate IN parameters at each IN layer in the network to shift back feature maps, however an important question remains. IN layers are mainly used in style transfer networks, and only IBN has explored the use of IN layers for standard visual recognition tasks~\cite{ibn}, which significantly limits the applicability of the lightweight de-stylization module to state-of-the-art architectures. Fortunately, modern networks, differing in the number of layers and types of layers used, are similar to the seminal AlexNet~\cite{krizhevsky2012imagenet} in design. They contain five blocks of convolution layers, topped by several fully connected layers for classification; and the blocks are different among different architectures. For example, ResNet50 contains BottleNeck blocks while Denseblocks in DenseNet~\cite{huang2017densely}. As a result, we can plug the output head of the de-stylization module after these convolutional blocks. We denote $\vv_k$ as the feature maps from the $k$-th block in a network, and before it is sent to the next block, we perform the normalization with Eqn.~\ref{eqn:head}, the outputs are defined as $g_k(\vv_k)$. Doing this directly might destroy the feature maps in original networks, which are optimized without any IN layers. To mitigate this issue, we further shortcut these layers with skip connections, and now features sent to the next block become:
\begin{equation}
\label{eqn:skip}
\vv^*_k = g_k(\vv_k) + \vv_k
\end{equation}
In this case, if the learned affine parameters are set to zero, then there is simply no normalization. And the network will degrade to the original network. We name the proposed extension as generalized de-stylization (\gsystem), illustrated in Fig.~\ref{fig:genIN}.

% the performance of the transformed network will be lower-bounded by the original network.

\textbf{Discussions}. The design of extending the de-stylization module with skip connections allows the model to remove style information in feature maps brought by applied filters and at the same time without hurting originally optimized features. Consequentially, we can simply optimize weights in the de-stylization module without re-training the base network. With less than $ 10\%$ parameters compared to an entire network needed for finetuning, the module can be learned efficiently.

\section{Experiments}
We first study the the robustness of pretrained networks on \instagram, and we get the upper bound performance by finetuning on these filtered images since their labels are readily available. Then we show the effectiveness of proposed \system and its generalized version \gsystem. After that, we show the performance when only a limited number of class samples are available, followed by some ablation studies and qualitative analysis.

\textbf{Implementation details}.
For both training and finetuning, we set the number of epochs to be 15 and use SGD as the optimizer. The learning rate is set to 0.001 and is decreased by a factor of 10 after 10 epochs. All experiments are conducted on NVIDIA 4 Quadro P6000 GPUs. We use a VGG pretrained on \imagenet as the encoder of \system, and we fixed its weights for all experiments. To test the performance of \system, we use an IBN with a pretrained ResNet50 as its backbone since it is the only network that contains IN layers used for recognition tasks. We compute Top1/Top5 accuracy for each type of filters in \instagram and report the mean accuracies across all filters. For \gsystem, we only perform normalization at the end of \texttt{Conv1} and \texttt{Conv2}, as feature divergence caused by appearance changes is large in these layers.

\textbf{Robustness of pretrained networks}. Table.~\ref{tab:drop} presents the results of a pretrained ResNet50 and IBN model when applied to \instagram. We can see significant performance drops for both ResNet50 and IBN. For example, Top1 accuracy dropped by 8.92 and 7.89 (absolute percentage) separately.  The degradation in accuraries of IBN is less severe than ResNet50, confirming the fact that IN layers can indeed help nomalize style information.

\begin{table}[h!]
\begin{center}
\resizebox{\linewidth}{!}{
\begin{tabular}{@{}*{7}c@{}}
\toprule
&& \imagenet  && \instagram \\
 \cmidrule{3-3} \cmidrule{5-5}
Method   && Top1/Top5 acc && Top1/Top5 acc  \\
\cmidrule{1-1} \cmidrule{3-3} \cmidrule{5-5}
Resnet50& & 76.13/ 92.93 && 67.21/ 87.62 \\
\cmidrule{1-1} \cmidrule{3-3} \cmidrule{5-5}

IBN && \textbf{77.44}/\textbf{93.69} && \textbf{69.55}/ \textbf{89.32} \\
\bottomrule
\end{tabular}}
\end{center}
\vspace{-10pt}
\caption{Performance of pretrained models on \instagram. Notice the significant performance drop.}
\label{tab:drop}
% \vspace{-10pt}
\end{table}

\textbf{Upper bound by finetuning}. Since the semantics are preserved after applying filters, we finetune both ResNet50 and IBN on \instagram with images in different types of filters together, denoted as ResNet50-ft and IBN-ft. Finally, we finetune ResNet50 for each filter type separately and refer this method as finetuning upperbound (UB). The results are summarized in Table.~\ref{tab:all}

It is worth mentioning that training models with all data is extremely computationally expensive and time consuming (\instagram is $20\times$ size of \imagenet). Besides, it also requires lots of space to store the data. Thus, we randomly select 10\% of images from each object category in the \imagenet training set, and transform each image with a random Instagram filter. As a result, we generated a mini version of \instagram training set, which we named as \instagram-mini. \instagram-mini is only $1/200$ size of \instagram. There are only around 6 images per object category for each Instagram filter type. We finetune ResNet50 and IBN using \instagram-mini and show the results in the third column of Table.~\ref{tab:all}

\begin{table}[h!]
\begin{center}
\resizebox{\linewidth}{!}{
\begin{tabular}{@{}*{7}c@{}}
\toprule
  && \instagram  && mini \\
   \cmidrule{3-3} \cmidrule{5-5}

  Method   && Top1/Top5 acc && Top1/Top5 acc  \\
 \cmidrule{1-1} \cmidrule{3-3} \cmidrule{5-5}
ResNet50-ft && 74.52/ 92.08 && 72.53/ 91.02 \\
IBN-ft && 75.47/ 92.71 && 73.23/91.53 \\
\cmidrule{1-1} \cmidrule{3-3} \cmidrule{5-5}
ResNet50-UB && 75.62/ 92.64 && -/ - \\
\bottomrule
\end{tabular}}
\end{center}
\vspace{-10pt}
\caption{Results of pretrained models \textit{vs.}  finetuned models on \instagram and \instagram-mini ($1/200$ size of \instagram).}    
\label{tab:all}
% \vspace{-10pt}
\end{table}

We can see with the entire \instagram training set, simply finetuning can improve the Top1 accuracy 
from 67.21\% to 74.52\% and from 69.55\% to 75.47\% for ResNet50 and IBN, separately. % but there is a still a $2\%$ gap compared to the performance of the original model on \imagenet. 
Besides, by comparing the performances of ResNet50-ft and ResNet50-UB, we can see finetuning together is almost as good as finetuning separately, but finetuning separately is less practical as it requires 20 separate models. Furthermore, with \instagram-mini as the training set, the Top1 accuracy of ResNet50 and IBN could improve by $5.32\%$ and $3.68\%$ compared to pretrained models.
Although the Top1 accuracy is still $2\%$ less than finetuning with the entire \instagram, it is much more computationally feasible. Therefore, we report results on \instagram-mini instead of the entire \instagram in the remaining of the paper.

\textbf{Effectiveness of \system}.
We evaluate the effectiveness of \system when plugged into an IBN on \instagram-mini. In addition to comparing with pretrained ResNet50 and IBN, we also compare with IBN-IN, in which only IN parameters are finetuned while other weights fixed. 
Besides the mean accuracies on all 20 filters on \instagram, we also show the averaged performance on 10 filters that lead to the most significant drops of ResNet50, named as Hard 10.
% The second group methods are methods with parameters unfixed: e)~ResNet50 and f)~IBN finetuning all parameters, and finally g)~adaptive-IBN finetuning all parameters. 
The results are shown in Table.~\ref{tab:10p}. We observe that the proposed \system achieves the best performance. On \instagram-mini, IBN-IN improves the Top 1 accuracy of IBN by $1.2\%$, which demonstrates that simply finetuning IN can help generalization slightly. \system further improves the performance of IBN-IN by $1.2\%$. The effectiveness of \system is clearly visible in the subset of 10 hard filters, where \system improves by $5\%$ over Top-1 accuracy compared with IBN.
% Especially when the original model are fixed, with instIN the adaptive IBN can improve about 2.5\% top1 accuracy in all 20 filter types. On the hard 10 filter types, the adaptive IBN improved by over 5\% top1 accuracy!

\begin{table}[h]
\begin{center}
\resizebox{0.75\linewidth}{!}{
\begin{tabular}{@{}*{3}c@{}}
\toprule
  & All 20  & Hard 10 \\
  Method   & Top1/Top5 acc & Top1/Top5 acc  \\
\midrule
ResNet50 & 67.21/ 87.62 & 62.41/ 84.471 \\
IBN & 69.55/ 89.32 & 65.19/ 86.71 \\
\midrule
IBN-IN & 70.73/ 90.08 & 67.68/ 88.39 \\
\system & \textbf{71.96}/ \textbf{90.93} & \textbf{70.31}/ \textbf{90.06} \\
% \midrule
% ResNet50-ft & 72.53/ 91.02 & 71.56/ 90.53 \\
% IBN-ft & 73.23/ 91.53 & 71.65/ 90.79 \\
% \midrule
% \system-ft & \textbf{73.53}/ \textbf{91.63} & \textbf{72.67}/ \textbf{91.18} \\
\bottomrule
\end{tabular}}
\end{center}
\vspace{-10pt}
\caption{Performances of \system on \instagram-mini. Our proposed \system achieves best results. The improvement on Hard 10 filters is significant.}    
\label{tab:10p}
% \vspace{-10pt}
\end{table}

\textbf{Effectiveness of \gsystem}.
We now investigate if we can extend \gsystem to modern architectures, by using three top-performing CNNs, \textit{i.e.}, ResNet50, DenseNet121~\cite{huang2017densely} and SEResNet50~\cite{hu2018squeeze}, whose Top1 accuracies on  \imagenet are $76.13\%$, $74.47\%$ and $77.61\%$, respectively. We apply the \gsystem at the end of Block1 and Block2, as shown in Figure.~\ref{fig:genIN}. To be specific, the applied positions are the end of \texttt{Conv2\_3} and \texttt{Conv3\_4} in ResNet50, the end of DenseBlock1 and DenseBlock2 in DenseNet121, and the end of \texttt{Conv2\_3} and \texttt{Conv3\_4} in SeResNet50. 

\begin{figure}[h!]
\includegraphics[width=\linewidth]{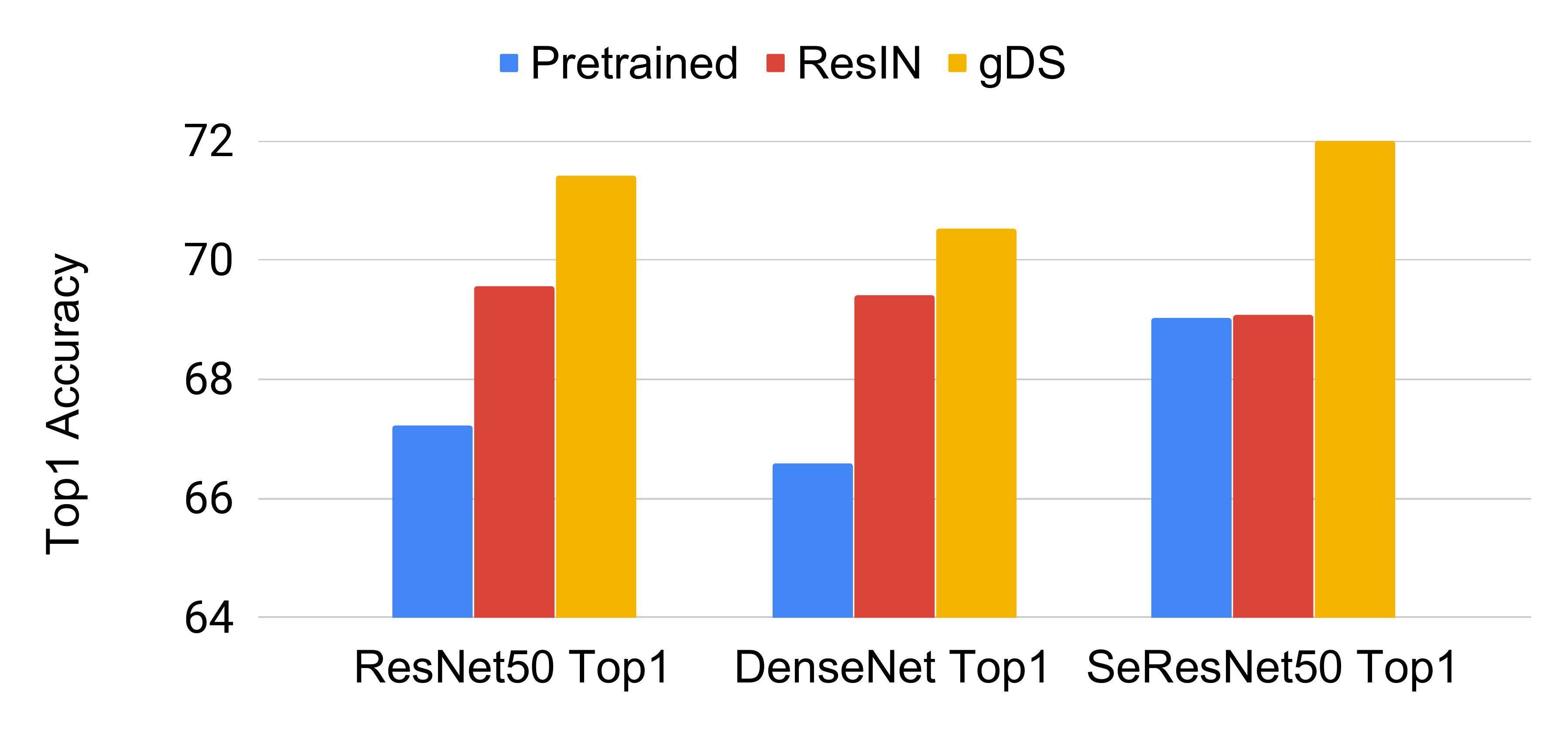}
\vspace{-15pt}
\caption{Performances of proposed \gsystem on different CNN architectures. We compare pretrained model, Finetuning ResIN and  \gsystem for each CNN architecture.}
\label{fig:gen}
\end{figure}

In addition to showing the results of pretrained networks and \gsystem, we also consider a simpler form of \gsystem---ResIN---which adds IN layers shortcut by skip connections at the same position of \gsystem to ResNet50. As in \gsystem, we only train the parameters of IN layers while keeping the remaining weights of the network fixed. The difference between ResIN and \gsystem is that, affine parameters of IN layers are conditioned on additional information. The results are summarized in Figure~\ref{fig:gen}.

We can see that, \gsystem outperforms pretrained models by ~$3\%-4\%$ for all architectures, which verifies that \gsystem can be applied to modern networks that help the generalization on filtered images. Further, both ResIN and \gsystem improve the performance of pretrained models, confirming the fact that normalizing feature maps with IN layers is helpful. As \gsystem normalizes feature maps conditioned on style information in feature maps from input samples, it helps learning separate parameters for different filters. Interestingly, SeResNet50 didn't benefit from ResIN much, but still benefits from \gsystem by $3\%$.

We also evaluate the effectiveness of the designed skip-connection. To achieve this, we remove the skip-connection in ResIN and finetune the IN parameters. The resulting Top1 accuracy is only 36.16\%, much worse than pretrained ResNet50. This indicates that without the skip-connection, the added normalization layer could even destroy the optimized feature.

\begin{figure}[ht!]
\centering
\includegraphics[width=1\linewidth]{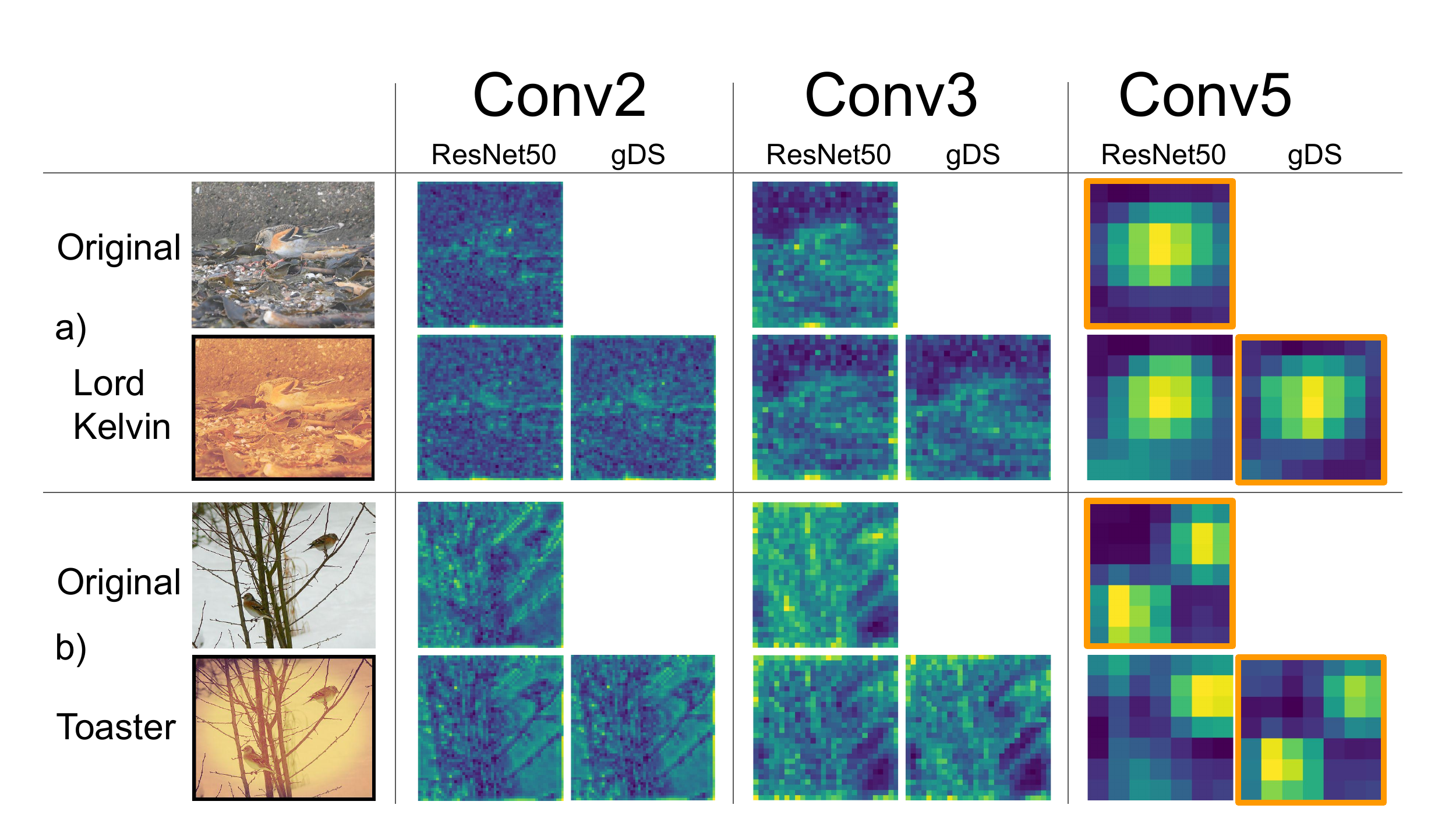}
\vspace{-5pt}
\caption{Feature maps comparison of ResNet50 and ResNet50+\gsystem. Two cases from two most challenging Instagram filers are shown. (a) ``Lord Kelvin'', (b)  ``Toaster''. For each example case, we show the original \imagenet image  and its Instagram-filtered version. For each image, we also show the activation feature maps after \texttt{Conv2}, \texttt{Conv3} and \texttt{Conv5}, generated by ResNet50 and Resnet50+\gsystem. \gsystem succesfully transforms the features of filtered images to be close to original images (See the diagonal comparisons).}
\label{fig:qual}
\end{figure}

The visualization of how \gsystem removed the style information brought by filters is shown in Fig.~\ref{fig:qual}. 

\textbf{Generalization to unseen classes}.
\label{sec: partial}
We demonstrate parameters used for scaling and shifting learned by \gsystem can generalize to unseen classes. During training, we randomly select half of the categories from \instagram-mini and use them to train the \gsystem. At test time, we evaluate the learned model on the other half of categories that have never been seen before from \instagram validation set, and on all categories as well. We compare with a pretrained ResNet50 and ResNet50-ft under the same settings and the results are shown in Table.~\ref{tab:partial}.

We can see that on unseen categories, the performance of ResNet50 finetuned using filtered images on seen categories decreases, while \gsystem increases the performance by $4.24\%$ compared to ResNet50. This demonstrates that \gsystem not only improves the generalization of CNNs on filtered images but also across different categories. When tested on all categories, finetuned ResNet50 improves upon ResNet50 slightly, but still worse than \gsystem.

\begin{table}[h!]
\begin{center}
\footnotesize
\resizebox{0.8\linewidth}{!}{
\begin{tabular}{@{}*{3}c@{}}
\toprule
 & Other Half & All \\
  Method   & Top1/Top5 acc & Top1/Top5 acc \\
\midrule
ResNet50 & 67.64/ 87.85 & 67.25/ 87.67 \\
\midrule
ResNet50 ft & 59.96/ 84.67 & 68.89/ 89.10 \\
\midrule
\gsystem & \textbf{71.40}/ \textbf{90.49} & \textbf{71.08}/ \textbf{90.30} \\
\bottomrule
\end{tabular}}
\end{center}\vspace{-10pt}
\caption{Performance of models learned with only half of the object categories, and evaluated on the other half of the object categories as well as all categories. }    
\label{tab:partial}
% \vspace{-10pt}
\end{table}

% \begin{table}[bt]
% \begin{center}
% \footnotesize
% \resizebox{0.5\linewidth}{!}{
% \begin{tabular}{@{}*{3}c@{}}
% \toprule
%  & Other Half & All \\
%   Method   & Top1/Top5 acc & Top1/Top5 acc \\
% \midrule
% ResNet50 & 67.64/ 87.85  \\
% \midrule
% ResNet50 ft & 59.96/ 84.67  \\
% \midrule
% \gsystem & \textbf{71.40}/ \textbf{90.49}  \\
% \bottomrule
% \end{tabular}}
% \end{center}
% \caption{Performance of models learned with only half of the object categories, and evaluated on the other half of the object categories. }    
% \label{tab:partial}
% % \vspace{-10pt}
% \end{table}

% \textbf{Number of training samples}.
% We investigate how the performance of \gsystem will change with respect to different number of samples used. We use a ResNet50 pretrained model and train the \gsystem with 2\%, 10\% (\instagram-mini) and 20\% of \instagram. The results are shown in Figure.~\ref{fig:pct}
% \begin{figure}[h!]
% \includegraphics[width=0.8\linewidth]{19ICCV_instagram/Figures/percent.pdf}
% \vspace{-15pt}
% \caption{Performance of \gsystem with different training data percentage.}
% \label{fig:pct}
% \end{figure}
% The results shown that the \gsystem could improve the pretrained ResNet50 peformance by \~4\%, even with 2\% of \instagram data. The averaged number of training image per category per Instagram filter is only 1.3.

\textbf{Comparisons with alternative methods}.
We compare \gsystem with two alternative methods based on ResNet50, \textit{i.e.}, AdaBN~\cite{li2016revisiting} and AdaIN~\cite{huang2017adain} that perform alignment without retraining the weights of the whole network. AdaBN~\cite{li2016revisiting} accumulates BN layer statistics on the target domain, without accessing training samples from \instagram. However, since AdaBN is designed for a single target domain, we apply AdaBN on the validation set of each filter type separately. 
In this way, AdaBN targets each Instagram filter type at a time, thus obtaining better performance than applying AdaBN on entire \instagram. On the other hand, AdaIN performs style transfer on filtered images such that the styles generated by filters are removed by an image generator. More specifically, we randomly select 100 images per Instagram filter from \instagram training set and 200 images from \imagenet to train a generator. Then the generator is used to sysnthesize filter-free images, upon which the original pretrained CNN is applied. The results are shown in Table.~\ref{tab:compare}.

\begin{table}[h!]
\begin{center}
\resizebox{0.8\linewidth}{!}{
\begin{tabular}{@{}*{2}c@{}}
\toprule
  Method   & Top1/Top5 acc   \\
\midrule
ResNet50 & 67.21/ 87.62  \\
\midrule
AdaBN~\cite{li2016revisiting} & 68.87/ 88,74  \\
\midrule
AdaIN~\cite{huang2017adain} & 35.45/ 58.86  \\
\midrule
\gsystem & \textbf{71.41}/ \textbf{90.46}  \\
\bottomrule
\end{tabular}}
\end{center}
\vspace{-10pt}
\caption{Comparison with other alternative methods on \instagram-mini. }    
\label{tab:compare}
% \vspace{-10pt}
\end{table}

We can see AdaBN improves the performance of pretrained models by only $1\%$ where as \gsystem obtained a $4.2\%$ performance gain. The results of AdaIN are worse than directly applying ResNet50. A reason could be that the synthesis process with an image generator is far from perfect and further introduces artifacts and distribution shifts.

\section{Conclusion}
We presented a study on how popular filters that are prevalent on social media affect the performance of pretrained modern CNN models. We created \instagram, by applying 20 pre-defined ImageNet filters to each image in \imagenet. We found that filters induce significant differences in feature maps compared to those of original images, and further lead to significant drops when directly applying CNNs pretrained on \imagenet. To improve generalization, we introduced a lightweight de-stylization module, which produces parameters used to scale and shift feature maps in order to recover changes brought by filters. Combining the lightweight module together with skip connections, we presented \gsystem that can be plugged into modern CNN networks. Extensive studies are conducted on \instagram and the results confirm the effectiveness of the proposed method.

\section{Acknowlegement}
This research was funded by ARO Grant W911NF1610342. 

\bibliographystyle{aaai}
\bibliography{reference}

\end{document}